\documentclass[sigconf, screen]{acmart}
\usepackage{booktabs}   
\usepackage{multirow}
\usepackage{pifont}
\usepackage{enumitem}
\AtBeginDocument{%
  }

\setcopyright{acmlicensed}
\copyrightyear{2026}
\acmYear{2026}
\acmDOI{XXXXXXX.XXXXXXX}
\acmConference{xx}{xx}{xx}
\acmISBN{978-1-4503-XXXX-X/2018/06}

\acmSubmissionID{3398}

\begin{document}

\title{Ground4D: Spatially-Grounded Feedforward 4D Reconstruction for Unstructured Off-Road Scenes}



\author{Shuo Wang}
\email{wangshuo@ict.ac.cn}
\affiliation{%
  \institution{Institute of Computing Technology, Chinese Academy of Sciences}
  \country{China}
}

\author{Jilin Mei}
\affiliation{%
  \institution{Institute of Computing Technology, Chinese Academy of Sciences}
  \country{China}
}

\author{Fuyang Liu}
\affiliation{%
  \institution{Institute of Computing Technology, Chinese Academy of Sciences}
  \country{China}
}

\author{Wenfei Guan}
\affiliation{%
  \institution{Institute of Computing Technology, Chinese Academy of Sciences}
  \country{China}
}

\author{Fanjie Kong}
\affiliation{%
  \institution{Xi'an Jiaotong University}
  \country{China}
}

\author{Zhihua Zhao}
\affiliation{%
  \institution{Beijing Institute of Technology}
  \country{China}
}

\author{Shuai Wang}
\affiliation{%
  \institution{Institute of Computing Technology, Chinese Academy of Sciences}
  \country{China}
}

\author{Chen Min}
\affiliation{%
  \institution{Institute of Computing Technology, Chinese Academy of Sciences}
  \country{China}
}

\author{Yu Hu}
\affiliation{%
  \institution{Institute of Computing Technology, Chinese Academy of Sciences}
  \country{China}
}

\renewcommand{\shortauthors}{wang et al.}
\renewcommand\footnotetextcopyrightpermission[1]{}
\settopmatter{printacmref=false} 

\begin{abstract}
Feedforward Gaussian Splatting has recently emerged as an efficient paradigm for 4D reconstruction in autonomous driving. However, in unstructured off-road scenes, its performance degrades due to high-frequency geometry, ego-motion jitter, and increased non-rigid dynamics. These factors introduce conflicting Gaussian observations across timestamps, leading to either over-smoothed renderings or structural artifacts. To address this issue, we propose Ground4D, a spatially-grounded 4D feedforward framework for pose-free off-road reconstruction. The key idea is to resolve temporal conflicts through spatially localized conditioning. Specifically, we introduce voxel-grounded temporal Gaussian aggregation, which partitions the canonical Gaussian space into spatial voxels and performs query-conditioned temporal attention within each voxel. Intra-voxel softmax normalization ensures that temporal selectivity and spatial occupancy become mutually reinforcing rather than conflicting. We furthermore introduce surface normal cues as auxiliary geometric guidance to regularize the geometry of Gaussian primitives. Extensive experiments on ORAD-3D and RELLIS-3D demonstrate that Ground4D consistently outperforms existing feedforward methods in reconstruction quality and generalizes zero-shot to unseen off-road domains. Project page and code: \url{https://github.com/wsnbws/Ground4D}
\end{abstract}



\begin{teaserfigure}
\centering
\includegraphics[width=0.99\textwidth]{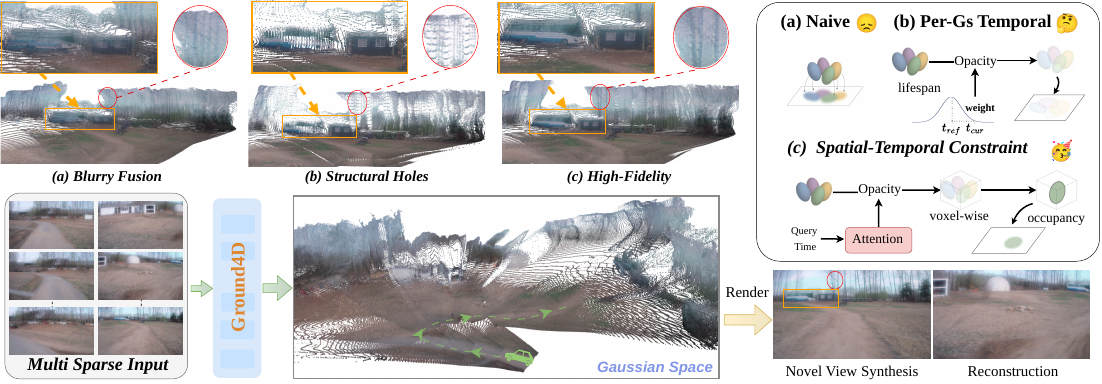}
\caption{Motivation for our work. Off-road scenes impose severe \textit{temporal relevance} demands on co-located Gaussians, observations from different timestamps that occupy the same canonical-space neighborhood. (a) Naive fusion (NopoSplat~\cite{ye2024no}) ignores temporal relevance, averaging conflicting attributes into blurry renderings. (b) Per-Gaussian temporal modeling (DGGT~\cite{chen2025dggt}) scores each Gaussian independently without spatial reference; when the query time is far from all context frames, opacity collapses into structural holes. (c) Ours resolves this via intra-voxel spatial grounding: confining temporal competition to each spatial voxel ensures that selectivity and occupancy are mutually reinforcing, achieving high-fidelity reconstruction.}
\Description{The figure is divided into upper and lower halves. The upper half shows three side-by-side 3D Gaussian point cloud reconstructions of an off-road scene with buildings, each viewed from the same angle. Version (a) labeled Blurry Fusion appears washed out with soft edges and red circles highlighting blurred regions. Version (b) labeled Structural Holes contains large black voids in the point cloud with red circles marking missing geometry. Version (c) labeled High-Fidelity shows a complete and sharp reconstruction without holes or blur. On the right side, three schematic diagrams explain the mechanisms: diagram (a) Naive shows multiple colored Gaussian ellipsoids merged directly into a single faded ellipsoid with an arrow to opacity. Diagram (b) Per-Gs Temporal shows ellipsoids scored by a lifespan curve where weights decay with temporal distance, leading to low opacity. Diagram (c) Spatial-Temporal Constraint shows ellipsoids grouped inside a voxel cube where attention produces voxel-wise occupancy, maintaining opacity. The lower half shows the full Ground4D pipeline: multi sparse input images feed into the system, producing a dense Gaussian space point cloud, which then renders a novel view and a reconstruction side by side.}
\label{fig:teaser}
\end{teaserfigure}


\maketitle

\section{Introduction}
Simulation environments are indispensable for the scalable training and evaluation of autonomous driving systems. Recent advances in neural scene representations~\cite{mildenhall2021nerf, kerbl20233d} have progressed from static reconstruction~\cite{muller2022instant, barron2022mip, barron2023zip, lu2024scaffold} to modeling complex dynamic 4D driving scenes~\cite{yang2023emernerf, zhou2024drivinggaussian, yan2024street, wu2023mars, chen2024omnire, huang2024textit, wei2025omni}. However, the reliance of these methods on per-scene optimization remains a bottleneck for scalable deployment. To address this, Feedforward Gaussian Splatting (FFGS)~\cite{smart2024splatt3r, lyu20243dgsr, chen2024pref3r, ye2024no, chen2025feat2gs, jiang2025anysplat} has emerged as a promising paradigm, leveraging 3D foundation models~\cite{wang2025vggt, shen2025fastvggt, wang2024dust3r, zhang2024monst3r, leroy2024grounding, cabon2025must3r, tang2025mv} to predict camera poses and dense geometry, and lifting pixels across time into a canonical Gaussian space without scene-specific training. While FFGS has shown strong results in structured urban settings~\cite{chen2025dggt}, its efficacy in unstructured off-road scenes remains largely unexplored.

Off-road autonomy, essential for agriculture, mining, and disaster response, demands reconstruction methods that go beyond the structured-scene assumption. Unlike urban scenes characterized by regular geometry, discrete rigid actors, and smooth ego-motion, off-road environments exhibit three \emph{mutually amplifying} properties: \textit{high-frequency geometry} from dense vegetation and loose terrain, \textit{spatially diffuse non-rigid dynamics} from swaying canopies and drifting particulates, and \textit{continuous ego-motion jitter} induced by rugged ground. High-frequency geometry amplifies jitter sensitivity, misaligning fine-grained observations, while diffuse dynamics further inject conflicting appearance into these misaligned signals, making existing FFGS methods built on structured scene, fundamentally inadequate.

These compounding factors manifest, in the canonical Gaussian space, as \textbf{attribute conflicts among co-located Gaussians}. Spatial-offset conflicts arise from ego-motion jitter that scatters surface observations, while temporal-variation conflicts stem from diffuse dynamics that produce inconsistent deformation states at fixed locations. As shown in Figure~\ref{fig:teaser}, current methods exhibit two failure modes under these conflicts. Naive fusion~\cite{ye2024no} blurs attributes, whereas per-Gaussian temporal modeling~\cite{chen2025dggt} collapses spatial occupancy into structural holes. Both failures share a common root. Temporal relevance is resolved independently of spatial context, forcing selectivity and occupancy into an inherent trade-off. Our key insight is that confining temporal competition to local spatial neighborhoods eliminates this trade-off. Within each neighborhood the locally dominant Gaussian simultaneously achieves the highest temporal relevance and guarantees spatial occupancy, making the two objectives mutually reinforcing.

Building on this insight, we present \textbf{Ground4D}, a spatially-grounded 4D feedforward reconstruction framework for pose-free off-road scenes. Given a set of context frames captured at different timestamps, Ground4D leverages the VGGT~\cite{wang2025vggt} backbone to predict camera parameters and depth, while introducing task-specific heads for Gaussian attributes, dynamic confidence, and surface normals. These components jointly lift pixels into a shared canonical Gaussian space, forming an initial 4D representation. To resolve the dual attribute conflicts, we introduce \textbf{voxel-grounded temporal Gaussian aggregation}: the canonical space is partitioned into spatial voxels, within each of which a query-conditioned temporal attention mechanism scores co-located Gaussians by their relevance to the query time. An intra-voxel softmax normalization then enforces that every non-empty voxel produces a valid primitive, making temporal selectivity and spatial occupancy \emph{mutually reinforcing} rather than conflicting. Gaussian attributes are subsequently aggregated through semantics-aware estimators that respect their respective geometric structures. For explicitly dynamic objects, a pretrained TAPIP3D model~\cite{tapip3d} interpolates motion across time. We further introduce surface normal regularization, which leverages normal cues as geometric priors to regularize Gaussian orientations in photometrically ambiguous off-road regions.
We conduct extensive experiments by training and evaluating diverse FFGS baselines, both pose-supervised and pose-free, on ORAD-3D~\cite{min2025advancing}. Ground4D surpasses all baselines across all metrics, achieving up to a 1.48~dB PSNR gain on ORAD-3D, and further generalizes zero-shot to RELLIS-3D~\cite{jiang2021RELLIS}. In summary, our contributions are:

\begin{itemize}[leftmargin=*, nosep]
\item We present Ground4D, a spatially-grounded 4D feedforward framework for pose-free off-road scene reconstruction.

\item We propose voxel-grounded temporal Gaussian aggregation, which confines temporal competition within spatial voxels via query-conditioned attention and intra-voxel normalization, making temporal selectivity and spatial occupancy mutually reinforcing. We further introduce surface normal regularization to improve geometric consistency.

\item Extensive experiments on ORAD-3D and RELLIS-3D demonstrate state-of-the-art performance among feedforward methods, with strong zero-shot generalization to unseen off-road domains.
\end{itemize}

\section{Related Works}

\begin{figure*}[h]
\centering
\includegraphics[width=\textwidth]{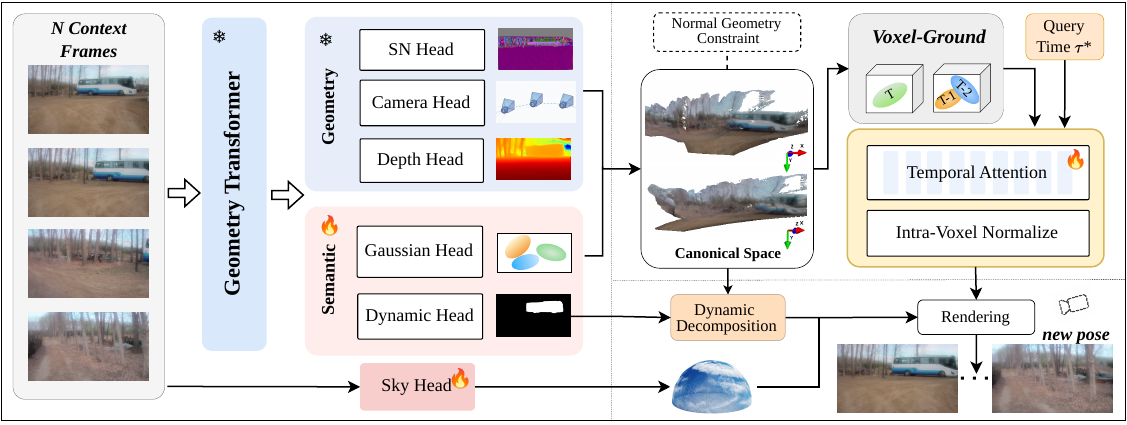}
\caption{Overview of our Ground4D framework. Ground4D transforms $N$ sparse context frames and a specific query time $\tau^*$ into a temporally-coherent 4D representation. Sparse multi-view observations are first unprojected into a shared canonical Gaussian space through a geometry-aware transformer\cite{wang2025vggt} to extract foundational scene primitives. These primitives then undergo voxel-grounded temporal aggregation, where query-conditioned attention mechanisms resolve spatial-temporal attribute conflicts within localized voxels. Finally, a joint rendering module composites the aggregated scene with interpolated dynamic Gaussians and a parametric sky model, enabling photorealistic synthesis at arbitrary viewpoints. SN Head represents surface normal head, whose intermediate features are injected into the Gaussian Head to provide geometric guidance.}
\Description{A pipeline diagram flowing left to right. On the left, N context frames of off-road scenes enter a frozen Geometry Transformer block. The transformer branches into two groups of prediction heads: a geometry group containing the SN Head producing a surface normal map, a Camera Head outputting camera poses, and a Depth Head generating a depth map; and a semantic group containing a Gaussian Head producing colored ellipsoids and a Dynamic Head outputting a binary mask. A dashed arrow connects the SN Head to the Gaussian Head indicating normal geometry constraint. The geometry and semantic outputs merge into a canonical Gaussian space visualized as a 3D point cloud. A downward arrow leads to a Dynamic Decomposition module. On the right side, a Voxel-Ground block receives a query time tau-star and the canonical space, showing colored voxel cubes with Gaussians grouped inside. Within this block, a Temporal Attention sub-module feeds into an Intra-Voxel Normalize sub-module. Below, a Sky Head with a trainable flame icon produces a sky dome. All three streams converge at a Rendering module that outputs novel-view images at a new camera pose. Snowflake icons mark frozen modules and flame icons mark trainable modules.}
\label{fig:ground4d}
\end{figure*}

\subsection{3D Reconstruction and View Synthesis}
3D scene reconstruction has transitioned from classical geometry pipelines~\cite{schonberger2016structure} to neural representations. NeRF~\cite{mildenhall2021nerf} pioneered this shift by optimizing continuous volumetric radiance fields, leading to extensions that improve anti-aliasing via conical frustums~\cite{barron2021mip}, handle unbounded environments through non-linear parameterization~\cite{barron2022mip}, and accommodate transient occlusions in unconstrained photo collections~\cite{martin2021nerf}. To mitigate NeRF's intensive computational overhead, subsequent works have leveraged sparse voxel grids~\cite{fridovich2022plenoxels}, tensor decomposition~\cite{chen2022tensorf}, and multiresolution hash encodings~\cite{muller2022instant} to accelerate training and rendering.

Alternatively, 3D Gaussian Splatting (3DGS)~\cite{kerbl20233d} introduces explicit anisotropic primitives for high-fidelity, real-time rendering. Recent advancements further refine its geometric consistency: Scaffold-GS~\cite{lu2024scaffold} employs voxel-anchored structures to reduce redundancy, while Mip-Splatting~\cite{yu2024mip} and 2DGS~\cite{huang20242d} introduce anti-aliasing filters and planar primitives, respectively, for geometrically accurate surface reconstruction. Functional extensions such as GaussianOpacity Fields~\cite{yu2024gaussian} facilitate precise mesh extraction, while 4DGS~\cite{wu20244d} and Deformable-GS~\cite{yang2024deformable} enable dynamic scene modeling via time-conditioned deformation. Despite these strides, both NeRF and 3DGS paradigms predominantly rely on per-scene optimization, necessitating the development of generalizable feedforward architectures.

\subsection{Feed-Forward Reconstruction}

Feedforward reconstruction methods amortize per-scene optimization by training across large multi-scene datasets, enabling single-pass inference on unseen scenes. Early NeRF-based generalizable methods, including PixelNeRF~\cite{yu2021pixelnerf}, IBRNet~\cite{wang2021ibrnet}, and MVSNeRF~\cite{chen2021mvsnerf}, condition radiance fields on pixel-aligned features or multi-view cost volumes, but inherit NeRF's slow volumetric rendering. Feedforward Gaussian methods offer a faster alternative. PixelSplat~\cite{charatan2024pixelsplat} regresses 3D Gaussians from image pairs via epipolar depth distributions. MVSplat~\cite{chen2024mvsplat} replaces this with plane-sweeping cost volumes, achieving $10\times$ fewer parameters and $2\times$ faster inference. DepthSplat~\cite{xu2025depthsplat} integrates monocular depth priors for robust large-scale multi-view reconstruction. NopoSplat~\cite{ye2024no} further removes the camera pose dependency, predicting Gaussians directly from uncalibrated inputs.

A deeper shift has emerged through 3D foundation models that jointly regress geometry and camera parameters. DUSt3R~\cite{wang2024dust3r} replaces the classical SfM pipeline with a transformer that jointly predicts point maps and poses from image pairs, while MASt3R~\cite{leroy2024grounding} augments it with dense feature matching for wide-baseline robustness. VGGT~\cite{wang2025vggt} marks a significant milestone. A single large transformer predicts camera parameters, depth, point maps, and 3D tracks from up to hundreds of images in one forward pass, consistently surpassing optimization-based alternatives. Building on VGGT's geometric priors, AnySplat~\cite{jiang2025anysplat} distills its representations into a feedforward Gaussian splatting model trained on nine diverse datasets, achieving strong zero-shot generalization across both sparse and dense view settings. VGGT-X~\cite{liu2025vggt} further scales VGGT inference to over one thousand images through memory-efficient implementation and global alignment refinement.

\subsection{Driving Scene Reconstruction}

Autonomous driving imposes unique challenges: unbounded scale, complex dynamic actors, sparse multi-camera setups, and simulation-grade fidelity. NeRF-based methods address these via LiDAR-RGB fusion~\cite{rematas2022urban}, compositional sub-networks for city-scale synthesis~\cite{tancik2022block}, and self-supervised flow fields for annotation-free static-dynamic decomposition~\cite{yang2023emernerf}. Scene graph formulations such as NSG~\cite{ost2021neural} and MARS~\cite{wu2023mars} attach local representations to individual actors, while UniSim~\cite{yang2023unisim} enables closed-loop sensor simulation with editable trajectories. The transition to 3DGS has significantly improved rendering efficiency. StreetGaussians~\cite{yan2024street} combines static backgrounds with tracked dynamic objects, S$^3$Gaussian~\cite{huang2024textit} achieves self-supervised decomposition through 4D consistency, and OmniRe~\cite{chen2024omnire} extends canonical Gaussian spaces to diverse dynamic agents via a unified scene graph. However, all these approaches rely on per-scene optimization, limiting scalability.

Recent efforts shift toward feedforward reconstruction for driving scenes. DrivingForward~\cite{tian2025drivingforward} enables fast Gaussian reconstruction from surround-view inputs via jointly trained pose, depth, and Gaussian networks. STORM~\cite{yang2024storm} directly predicts dynamic Gaussians with per-Gaussian velocity in a single forward pass, leveraging self-supervised scene flow aggregation, but is constrained by a fixed number of input frames. DrivingRecon~\cite{lu2024drivingrecon} improves generalization through large-scale pretraining with a prune-and-dilate strategy for multi-view fusion. Most recently, DGGT~\cite{chen2025dggt} introduces a pose-free formulation that jointly infers Gaussian representations and camera parameters from unposed inputs, employing a lifespan-based temporal visibility mechanism for dynamic handling. Despite this progress, existing feedforward driving methods are predominantly designed for structured urban environments with discrete rigid actors and smooth ego-motion. Off-road scenes, characterized by high-frequency geometry, remain unexplored, motivating the present work.

\section{Method}

\subsection{Problem Formulation}
Given a sequence of context frames $\mathcal{I} = \{I_t\}_{t=1}^{T}$, where $I_t \in \mathbb{R}^{H \times W \times 3}$ with normalized timestamps $\tau_t \in [0,1]$, we define the pixel domain $\Omega = \bigcup_{t=1}^{T} \Omega_t$, with $\Omega_t$ denoting the pixel of $I_t$. We seek a 4D scene representation that explains observations over $\Omega$ and renders a photorealistic image $\hat{I}$ at an arbitrary query time $\tau^{*}$ from any camera parameters $(E,K) \in \mathrm{SE}(3) \times \mathbb{R}^{3\times3}$. To this end, we adopt 3DGS~\cite{kerbl20233d} as the underlying representation, modeling the scene as a collection of Gaussians, where each primitive $\mathbf{g} = (\boldsymbol{\mu}, \mathbf{c}, \boldsymbol{\alpha}, \mathbf{s}, \mathbf{q})$ encodes mean, color, opacity, and a covariance parameterized by scale and rotation.

As illustrated in Figure~\ref{fig:ground4d}, Ground4D proceeds in three stages. \textbf{(1) Canonical Gaussian Space Construction} (Section~\ref{sec:backbone}): a frozen VGGT backbone jointly processes all $T$ frames. Task-specific heads predict camera parameters, depth, Gaussian attributes, surface normals, sky model and dynamic confidence, which are used to unproject pixels into a shared canonical space $\mathcal{G}^{s} = \bigcup_{t} \mathcal{G}_t^s$. \textbf{(2) Voxel-Grounded Temporal Gaussian Aggregation} (Section~\ref{sec:voxel}): the context Gaussians are partitioned into spatial voxels. Within each voxel, query-conditioned temporal attention scores co-located Gaussians by relevance to query time, and intra-voxel softmax normalization enforces temporal selectivity and spatial occupancy simultaneously, and the weighted Gaussians are fused into a single primitive per voxel. \textbf{(3) Rendering}: the refined Gaussians are composited with interpolated explicit dynamic Gaussians and a parametric sky background, then rasterized via 3DGS splat~\cite{kerbl20233d} for specified camera parameter and time to render the final image.

\subsection{Canonical Gaussian Space Construction}
\label{sec:backbone}
Scene understanding from an uncalibrated off-road context frames requires simultaneous estimation of camera parameters, dense depth, and per-pixel material properties. We build on VGGT~\cite{wang2025vggt}, a large geometric vision transformer that processes all $T$ context frames jointly through an alternating-attention aggregator, producing three complementary token streams: aggregated cross-view tokens $\mathbf{F}^{\text{agg}}_t$, image-level tokens $\mathbf{F}^{\text{img}}_t$, and DINOv2-derived semantic tokens $\mathbf{F}^{\text{dino}}_t$. The backbone is kept \emph{frozen} during fine-tuning; only the prediction heads and temporal fusion module are trained.

\textbf{Canonical Points Estimation.} A camera head decodes $\mathbf{F}^{\text{agg}}_t$ into a compact pose encoding from which extrinsics $E_t$ and intrinsics $K_t$ are recovered. A DPT\cite{ranftl2021vision} depth head produces per-pixel depth $D_t$. Then, 3D points of each pixel $(u,v)\in\Omega_t$ are obtained by unprojection:
\begin{equation}
  \mathcal{P}_t(u,v)
  = R_t^{\top}\bigl(K_t^{-1}\,D_t(u,v)\,[u,v,1]^{\top} - T_t\bigr),
\end{equation}
where $E_t = (R_t, T_t)$ denotes the camera extrinsics, with $R_t$ and $T_t$ being the rotation and translation defined with respect to the reference frame (taken as the first frame). Each unprojected point $\mathcal{P}_t(u,v)$ serves as the mean $\boldsymbol{\mu}$ of the corresponding Gaussian primitive, anchoring its spatial position in the canonical space.

\textbf{Gaussian Map.}
A DPT decoder operating on $\mathbf{F}^{\text{img}}_t$ and $\mathbf{F}^{\text{dino}}_t$  predicts a dense Gaussian attribute map:
\begin{equation}
  \Phi_t = \operatorname{GaussianHead}([\mathbf{F}^{\text{img}}_t;\,\mathbf{F}^{\text{dino}}_t])
  \;\in\; \mathbb{R}^{H \times W \times C_g},
\end{equation}
where $C_g = 11$ channels encode color, opacity, scale and rotation.

\textbf{Surface Normal Head.}
To inject explicit geometric priors into the Gaussian prediction, we introduce a normal head that decodes $\mathbf{F}^{\text{img}}_t$ and $\mathbf{F}^{\text{dino}}_t$  into a per-pixel unit normal map $\mathbf{n}_t$. Specifically, intermediate features from the normal head are injected into the Gaussian head, directly conditioning the prediction of Gaussian orientation and scale on the underlying surface geometry:
\begin{equation}
  \mathbf{n}_t = \operatorname{NormalHead}([\mathbf{F}^{\text{img}}_t;\,\mathbf{F}^{\text{dino}}_t])
  \;\in\; \mathbb{R}^{H \times W \times 3},
\end{equation}
where each pixel is assigned a 3D unit normal vector in camera coordinates, $\ell_2$-normalized to unit length. The normal head is additionally supervised against ground-truth normals to provide strong geometric constraints. The corresponding loss terms are detailed in Section~\ref{sec:loss}.

\textbf{Explicit Dynamic Gaussian Decomposition.}
Although explicitly dynamic objects such as vehicles and pedestrians are rare in off-road sequences, their occasional presence must be handled. Following DGGT~\cite{chen2025dggt}, a dynamic confidence head produces a per-pixel dynamic score $\Pi_t^{\text{dyn}} \in [0,1]$. Each Gaussian opacity in Gaussian map is weighted by $(1 - \Pi_t^{\text{dyn}})$ to suppress dynamic regions. For the dynamic Gaussians, we use TAPIP3D\cite{tapip3d} to track sparse 3D points on dynamic regions between keyframes, and linearly interpolate dynamic Gaussian centers to query times before compositing them with the refined Gaussians. However, the pervasive high-frequency non-rigid deformations cannot be resolved by this way and are instead handled implicitly by the voxel-grounded temporal attention aggregation described in Section~\ref{sec:voxel}.

\textbf{Sky Background Model.}
Open-sky regions are geometrically ill-defined and cannot be bounded within a 3D Gaussian field. A parametric sky model $\mathcal{M}^{\text{sky}}$ regresses a background appearance conditioned on viewing direction. The final composite is:
\begin{equation}
  \hat{I}_t
  = \mathbf{A}_t \odot \hat{I}_t^{3D}
  + (1-\mathbf{A}_t) \odot \mathcal{M}^{\text{sky}}(\mathcal{I}, E_t, K_t),
\end{equation}
where $\mathbf{A}_t$ is the accumulated alpha from 3DGS rasterization.

\subsection{Voxel-Grounded Temporal Gaussian Aggregation}
\label{sec:voxel}
Upon constructing the canonical Gaussian space $\mathcal{G}^s$, it contains $N = \mathcal{O}(T \cdot H \cdot W)$ primitives, each encoding a single observation of the scene at timestamp $\tau_t$. We then proceed in three steps: differentiable spatial voxelization groups co-located Gaussians into spatial voxels; query-conditioned temporal attention computes the temporal relevance of each Gaussian with respect to the query time; and intra-voxel normalization and fusion aggregates the weighted Gaussians within each voxel to ensure structural completeness and enhance temporal selectivity.

\textbf{Spatial Voxelization.}
We partition the Gaussian space $\mathcal{G}^s$ into spatially coherent groups via a uniform voxel grid of cell side $\rho$. Each Gaussian center $\boldsymbol{\mu}_i$ is quantized to integer coordinates:
\begin{equation}
  \mathbf{v}_i = \operatorname{round}\!\left(\frac{\boldsymbol{\mu}_i}{\rho}\right) \in \mathbb{Z}^3,
\end{equation}
where each unique integer coordinate $\mathbf{v}_i$ defines a voxel. All Gaussians sharing the same coordinate are assigned to the same voxel, and we denote the index set of Gaussians within voxel $m$ as $\mathcal{V}_m$. This yields $M \ll N$ non-empty sparse voxels via exact deduplication, partitioning $\mathcal{G}^s = \bigcup_{m=1}^{M} \mathcal{V}_m$ into variable-size groups while preserving all primitive attributes, including per-frame timestamps. The voxelization step is non-differentiable and used solely for grouping. Gradients are propagated through the aggregated Gaussian attributes in the fusion stage below.

\textbf{Query-Conditioned Temporal Attention.}
For each voxel $m$, we compute a query-conditioned relevance score for every co-located Gaussian $g_i \in \mathcal{V}_m$. Each primitive carries a feature $\mathbf{f}_i^{\text{gs}}$, where the Gaussian attributes are first encoded into a feature embedding, which is then concatenated with the residual feature from the last layer of the Gaussian head. Then,  $\mathbf{f}_i^{\text{gs}}$is projected into a shared latent space through a feature encoder $\phi_f$. In parallel, the normalized context timestamp $\tau_i$ is embedded via a multi-frequency sinusoidal encoding $\gamma(\cdot)$~\cite{mildenhall2021nerf} and a time MLP $\phi_t$, capturing temporal dynamics across multiple scales. The two streams are additively fused into a unified context representation:
\begin{equation}
  \mathbf{h}_i = \phi_f\bigl(\mathbf{f}_i^{\text{gs}}\bigr) + \phi_t\bigl(\gamma(\tau_i)\bigr).
\end{equation}
The query time $\tau^*$ is encoded similarly, yielding $\mathbf{h}^* = \phi_t\!\bigl(\gamma(\tau^*)\bigr)$. A lightweight attention MLP then operates on the concatenation of the context and query representations to produce a scalar temporal relevance logit:
\begin{equation}
  a_i = \operatorname{MLP}_{\mathrm{attn}}\bigl([\mathbf{h}_i;\,\mathbf{h}^*]\bigr).
\end{equation}

\textbf{Intra-Voxel Normalization and Fusion.}
Raw logits are converted to normalized attention weights via a temperature-scaled softmax applied independently within each voxel:
\begin{equation}
  w_i = \frac{\exp(a_i/\beta)}{\sum_{j\in\mathcal{V}_m}\exp(a_j/\beta)},
\end{equation}
where $\mathcal{V}_m$ denotes the index set of all Gaussian primitives assigned to voxel $m$,
and $\beta$ is a temperature parameter.
Restricting the softmax normalization to within each voxel confers two complementary properties: (\emph{i}) \textit{temporal selectivity}: the locally dominant gaussian, i.e., the one most relevant to $\tau^*$, drives the fused output while suppressing conflicting observations; and (\emph{ii}) \textit{spatial occupancy}: since the normalization is 
confined to each voxel, every non-empty voxel is guaranteed to produce a valid primitive regardless of how distant $\tau^*$ is from all context frames, ensuring structural completeness.

The learned weights $w$ then drive a set of attribute-specific fusion estimators, each tailored to the geometric semantics of the corresponding Gaussian parameter. Position and color, being Euclidean quantities, admit a straightforward attention-weighted mean. Opacity requires special treatment. Naive averaging collapses density in transiently occluded voxels, so we blend the weighted mean with the per-voxel maximum via a fixed mixing coefficient $\lambda$. Scale carries multiplicative rather than additive semantics, so aggregation is performed in log-space to yield a weighted geometric mean. Rotation, defined on the non-Euclidean manifold of unit quaternions, is handled via a
weighted Euclidean mean followed by $\ell_2$ re-normalization:
\begin{equation}
\left\{
\begin{aligned}
    \hat{\boldsymbol{\mu}} &= \sum_{i\in\mathcal{V}_m} w_i\,\boldsymbol{\mu}_i~~~;
    \quad
    \hat{\mathbf{c}} = \sum_{i\in\mathcal{V}_m} w_i\,\mathbf{c}_i \\[6pt]
    \hat{\boldsymbol{\alpha}} &= \lambda\max_{i\in\mathcal{V}_m}\boldsymbol{\alpha_i}
                + (1{-}\lambda)\sum_{i\in\mathcal{V}_m}w_i\boldsymbol{\alpha_i} \\[6pt]
    \hat{\mathbf{s}} &= \exp\Bigl(\sum_{i\in\mathcal{V}_m}w_i\log\mathbf{s_i}\Bigr) \\[6pt]
    \hat{\mathbf{q}} &= \frac{\sum_{i\in\mathcal{V}_m}w_i\mathbf{q_i}}
                          {\bigl\|\sum_{i\in\mathcal{V}_m}w_i\mathbf{q_i}\bigr\|_2} \quad ,
\end{aligned}
\right.
\end{equation}
where all index sets are confined to $\mathcal{V}_m$, ensuring that each voxel is fused
independently without cross-voxel interference.

\textbf{Temporal Augmentation.}
To prevent the fusion network from degenerating into a nearest-neighbor lookup, we introduce random frame dropout: during training, the frame at query time $\tau^*$ is withheld from the voxelization space with probability $p_{\text{drop}}$, while its rendered output is retained as the photometric supervision target. This forces the network to infer the scene state at $\tau^*$ from temporally non-contiguous context, regularizing against overfitting to transient deformations and improving generalization across disjoint observation windows.

\subsection{Training Objectives}
\label{sec:loss}
We supervise the full pipeline end-to-end with a composite objective. The overall loss $\mathcal{L}$ is defined as the sum of its individual components:
\begin{equation}
  \mathcal{L} = \mathcal{L}_{\text{rgb}} + \mathcal{L}_{\text{lpips}} + \mathcal{L}_{\text{dyn}} + \mathcal{L}_{\text{sky}} + \mathcal{L}_{\text{norm}}^{\text{pred}} + \mathcal{L}_{\text{norm}}^{\text{gs}}.
\end{equation}
For photometric supervision, the primary signal is an $\ell_1$ loss between the rendered image $\hat{I}_t$ and the ground-truth $I_t$ across $T$ views:
\begin{equation}
  \mathcal{L}_{\text{rgb}} = \lambda_{\text{rgb}} \frac{1}{T}\sum_{t=1}^{T}\|I_t - \hat{I}_t\|_1.
\end{equation}
To recover high-frequency textural details that $\ell_1$ penalizes insufficiently, we incorporate an LPIPS perceptual loss~\cite{zhang2018unreasonable}:
\begin{equation}
  \mathcal{L}_{\text{lpips}} = \lambda_{\text{lpips}} \frac{1}{T}\sum_{t=1}^{T}\operatorname{LPIPS}(I_t,\hat{I}_t).
\end{equation}
To explicitly model dynamic elements and sky regions, we introduce mask-based supervision. The dynamic confidence head is trained with a binary cross-entropy loss against per-view dynamic annotations $\mathcal{M}_t^{\text{dyn}}$:
\begin{equation}
  \mathcal{L}_{\text{dyn}} = \lambda_{\text{dyn}} \frac{1}{T}\sum_{t=1}^{T} \operatorname{BCE} \bigl(\Pi_t^{\text{dyn}},\,\mathcal{M}_t^{\text{dyn}}\bigr).
\end{equation}
Similarly, to ensure unbounded sky regions are correctly delegated to the background model, the accumulated alpha $\mathbf{A}_t$ is supervised against the inverse sky mask ground truth $M_t^{\text{sky}}$:
\begin{equation}
  \mathcal{L}_{\text{sky}} = \lambda_{\text{sky}} \frac{1}{T}\sum_{t=1}^{T}\bigl\|\mathbf{A}_t - (1 - M_t^{\text{sky}})\bigr\|_1.
\end{equation}
\textbf{Surface Normal Loss.} Finally, we enforce geometric consistency through two surface normal losses evaluated on valid static pixels where ground-truth normals $\bar{\mathbf{n}}_t$ are available. First, a predicted-normal loss measures the cosine distance between the NormalHead output $\mathbf{n}_t$ and $\bar{\mathbf{n}}_t$:
\begin{equation}
  \mathcal{L}_{\text{norm}}^{\text{pred}} = \lambda_{\text{norm}}^{\text{pred}} \frac{1}{|{\Omega_{valid}}|} \sum_{(u,v)\in{\Omega_{valid}}} \Bigl(1 - \bigl|\langle\mathbf{n}_t(u,v),\,\bar{\mathbf{n}}_t(u,v)\rangle\bigr|\Bigr).
\end{equation}
where $\langle\cdot,\cdot\rangle$ denotes the inner product between two unit vectors.

Second, we introduce a gaussian-derived normal loss $\mathcal{L}_{\text{norm}}^{\text{gs}}$. Since each 3DGS ellipsoid implicitly defines a surface normal through its direction of minimum spatial extent, we estimate it via a soft-argmin over the three scale axes:
\begin{equation}
  \tilde{\mathbf{n}}
  = R_{\mathbf{q}}\,
    \frac{\operatorname{softmax}(-\eta\,\mathbf{s})}
        {\|\operatorname{softmax}(-\eta\,\mathbf{s})\|_2},
  \label{eq:gs_normal}
\end{equation}
where $R_{\mathbf{q}}$ is the Gaussian rotation matrix and $\eta$ is a temperature parameter. The axis with the smallest scale receives the highest weight, identifying the normal direction of the disc-like gaussian. This estimated normal is projected into camera coordinates and supervised against the ground truth:
\begin{equation}
  \mathcal{L}_{\text{norm}}^{\text{gs}} = \lambda_{\text{norm}}^{\text{gs}} \frac{1}{|{\Omega_{valid}}|} \sum_{(u,v)\in{\Omega_{valid}}} \Bigl(1 - \bigl|\langle R_{t}\,\tilde{\mathbf{n}}(u,v),\, \bar{\mathbf{n}}_t(u,v)\rangle\bigr|\Bigr).
\end{equation}
This coupled geometry-aware signal strongly regularizes the Gaussian orientations, aligning them with the true scene surface structure.
\begin{table*}[!t]
  \centering
    \caption{Comparison of different FFGS methods on off-road datasets. Bold indicates the best performance. Time and Memory are measured on a single A6000 GPU at $256 \times 448$ resolution with 4 context frames.}
  \label{tab:comparsion_table}
  \begin{tabular}{lcccccccccc}
    \toprule
    \multirow{2}{*}{Methods} 
    & \multicolumn{3}{c}{ORAD-3D~\cite{min2025advancing}} 
    & \multicolumn{3}{c}{RELLIS-3D~\cite{jiang2021RELLIS} (Zero-shot)} 
    & \multirow{2}{*}{Dynamic} 
    & \multirow{2}{*}{Pose-free} 
    & \multirow{2}{*}{Time (s)} 
    & \multirow{2}{*}{Memory (MB)} \\
    
    \cmidrule(r){2-4} \cmidrule(l){5-7}
    & PSNR $\uparrow$ & SSIM $\uparrow$ & LPIPS $\downarrow$ 
    & PSNR $\uparrow$ & SSIM $\uparrow$ & LPIPS $\downarrow$ \\
    
    \midrule
    MvSplat~\cite{chen2024mvsplat}    
    & 15.01 & 0.30 & 0.53 
    & 9.95  & 0.16 & 0.68 
    & \ding{56} & \ding{56} & 0.03 & 2370 \\

    DepthSplat~\cite{xu2025depthsplat}
    & 21.98 & 0.60 & 0.37 
    & \textbf{22.29} & 0.52 & \underline{0.34} 
    & \ding{56} & \ding{56} & 0.01 & 7708 \\

    STORM~\cite{yang2024storm}      
    & 20.56 & 0.54 & 0.44 
    & 18.40 & 0.46 & 0.56 
    & \ding{52} & \ding{52} & 0.08 & 16932 \\

    NopoSplat~\cite{ye2024no}  
    & \underline{22.41} & \underline{0.62} & 0.33 
    & 21.40 & 0.50 & 0.38 
    & \ding{56} & \ding{52} & 1.47 & 9744 \\

    DGGT~\cite{chen2025dggt}       
    & 21.76 & 0.61 & \underline{0.32} 
    & 21.27 & \underline{0.53} & 0.36 
    & \ding{52} & \ding{52} & 0.08 & 15576 \\

    \midrule
    \textbf{Ground4D (Ours)} 
    & \textbf{23.89} & \textbf{0.64} & \textbf{0.23} 
    & \underline{22.12} & \textbf{0.55} & \textbf{0.28} 
    & \ding{52} & \ding{52} & 0.17 & 15566 \\
    
    \bottomrule
  \end{tabular}
\end{table*}

\section{Experiment}
\subsection{Experimental protocol}
\textbf{Datasets.} We evaluate our model on ORAD-3D and RELLIS-3D. For training, we utilize the ORAD-3D training split, covering 103 sequences (300–600 frames each) across diverse off-road environments, including woodlands, grasslands, and rural roads. Evaluation is performed on 30 designated ORAD-3D test splits. Specifically, each split is partitioned into 70-frame clips, where 4 reference views are selected to reconstruct the remaining frames. To assess zero-shot generalization, we further evaluate on RELLIS-3D following the identical inference protocol.

\textbf{Implementation details.} We freeze the VGGT backbone and train only the Gaussian, dynamic, normal heads, the sky model, and the temporal fusion module. Images are resized to $518 \times 518$. Key hyperparameters: voxel size $\rho = 0.002$, temporal encoding with $10$ frequency bands and hidden dimension $64$, opacity mixing coefficient $\lambda = 0.3$, learnable temperature $\beta$ initialized to $1.0$, and softmin temperature $\eta = 10.0$ (Eq.~\ref{eq:gs_normal}). We use AdamW with cosine annealing ($1\text{K}$-step warmup) for $2\text{K}$ epochs, with peak learning rates of $1 \times 10^{-4}$ (sky and fusion) and $4 \times 10^{-5}$ (remaining heads). Training runs on two A6000 GPUs with batch size $1$, sampling 4-frame sequences with query-frame dropout probability $p_{\text{drop}} = 0.7$. Loss weights: $\lambda_{\text{rgb}} = 1.0$, $\lambda_{\text{sky}} = 1.0$, $\lambda_{\text{dyn}} = 0.5$, $\lambda_{\text{lpips}} = 0.05$, $\lambda_{\text{norm}}^{\text{pred}} = 0.05$, $\lambda_{\text{norm}}^{\text{gs}} = 0.02$. Ground-truth normals are derived geometrically by unprojecting calibrated dense depth into 3D and estimating per-pixel normals via multi-directional differential convolution, with sky and dynamic regions masked out.

\textbf{Evaluation Metrics.} Reconstruction performance is assessed via PSNR, SSIM, and LPIPS. This metric triad balances low-level signal fidelity with high-level perceptual quality, providing a holistic evaluation of the complex textures and adverse lighting conditions inherent in off-road terrains.

\subsection{Off-Road Novel-View Synthesis}
Table~\ref{tab:comparsion_table} compares Ground4D against five FFGS baselines on ORAD-3D and RELLIS-3D under identical settings.

\noindent\textbf{Results on ORAD-3D.} Ground4D achieves state-of-the-art performance across all metrics, with a 1.48~dB PSNR gain over the strongest baseline. Pose-supervised methods MvSplat and DepthSplat, despite having access to ground-truth camera parameters, suffer from noisy off-road poses and the absence of dynamic modeling, leading to degraded reconstruction fidelity. DGGT and STORM explicitly handle dynamics via instance masks and optical flow, yet both are designed for discrete rigid actors and prove inadequate for the pervasive high-frequency non-rigid deformations that characterize off-road scenes. NopoSplat achieves competitive pixel-level accuracy but exhibits poor perceptual quality, as its static scene assumption fails under the stochastic micro-dynamics prevalent in off-road environments. In contrast, Ground4D resolves temporal conflicts locally within spatial voxels, simultaneously suppressing attribute ambiguity and preserving spatial occupancy.

\noindent\textbf{Zero-Shot Generalization to RELLIS-3D.} Without any fine-tuning, Ground4D achieves the best SSIM and LPIPS among all evaluated methods and the highest PSNR among pose-free methods. DepthSplat attains a marginally higher PSNR by leveraging ground-truth poses, yet Ground4D surpasses it in both structural similarity and perceptual quality despite operating without any pose supervision. This indicates that intra-voxel temporal grounding learns a transferable scene prior. Even without access to calibrated poses, the spatially localized aggregation preserves geometric coherence that generalizes across diverse off-road domains.

\begin{table}[!h]
  \centering
  \caption{Generalization on urban driving scenes.}
  \label{tab:urban}
  \begin{tabular}{lcccc}
    \toprule
    Methods  & PSNR $\uparrow$ & SSIM $\uparrow$ & LPIPS $\downarrow$ \\
    \midrule
    MVSplat~\cite{chen2024mvsplat}     & 22.83 & 0.629 & 0.317          \\
    DepthSplat~\cite{xu2025depthsplat}  & 19.52 & 0.601 & 0.376          \\
    Drivingforward~\cite{tian2025drivingforward}  & 26.06 & 0.781 & 0.215          \\
    STORM~\cite{yang2024storm}  & 24.54 & 0.784 & 0.267          \\
    DGGT~\cite{chen2025dggt}  & \underline{26.63} & \underline{0.813} & \textbf{0.122} \\
    \midrule
    \textbf{Ours} & \textbf{27.23} & \textbf{0.814} & \underline{0.135} \\
    \bottomrule
  \end{tabular}
\end{table}

\subsection{Extension to Urban Scenes}

\begin{figure*}[t]
  \centering
  \includegraphics[width=\textwidth]{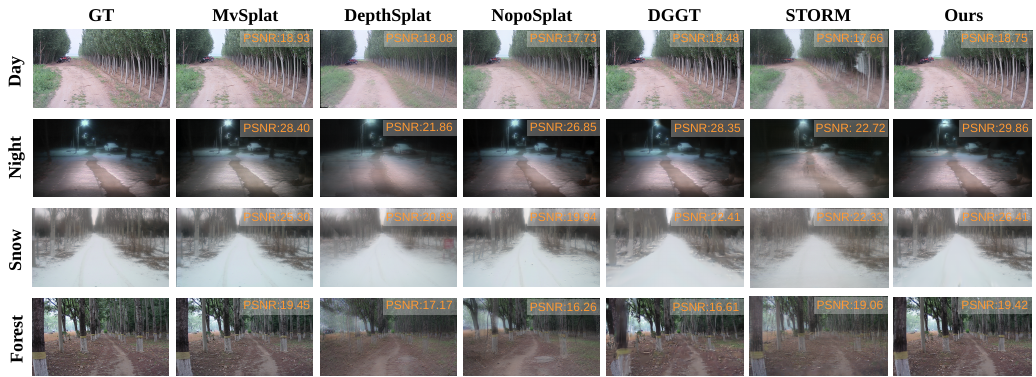}
  \caption{Reconstruction quality on input context frames of ORAD-3D dataset. We compare Ground4D against different baselines\cite{xu2025depthsplat, chen2024mvsplat, yang2024storm, ye2024no, chen2025dggt} on four off-road scenarios. Notably, all visualized frames are context frames used as input, not novel views. Ours achieves consistently clean reconstructions across all conditions.}
  \Description{A grid of reconstructed images comparing seven methods (GT, MvSplat, DepthSplat, NopoSplat, DGGT, STORM, and Ours) across four off-road scenarios (Day, Night, Snow, Forest). Each cell shows a rendered frame with its PSNR value overlaid in orange. Baseline methods exhibit visible artifacts: MvSplat and DepthSplat show geometric distortions, NopoSplat produces blurry outputs, DGGT contains semi-transparent holes, and STORM introduces surface warping. The Ground4D results in the rightmost column appear visually closest to the ground truth with the highest PSNR values across all four conditions.}
  \label{fig:contrast_vis}
\end{figure*}

\begin{figure*}[h]
  \centering
  \includegraphics[width=\textwidth]{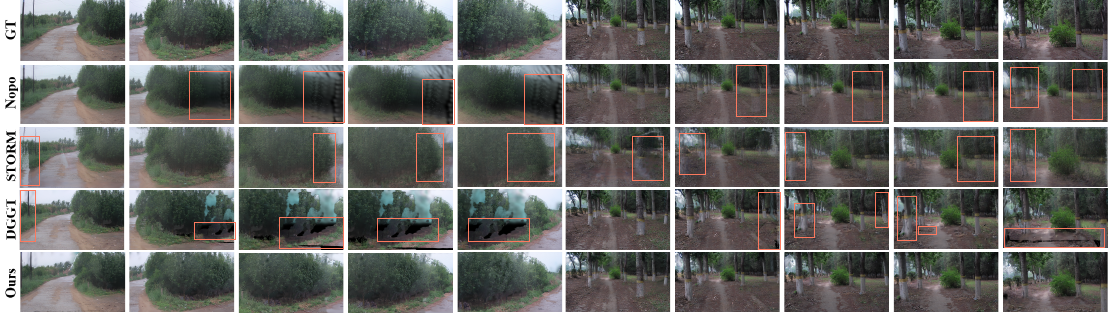}
  \caption{Novel-view synthesis consistency under off-road ego-motion. We show five consecutive synthesized frames for four pose-free methods across two challenging sequences. All visualized frames are held-out views not seen during reconstruction. Ours produces spatio-temporally consistent novel views with sharp details and complete structure throughout both sequences.}
  \Description{Two sets of five consecutive novel-view frames arranged in rows, comparing four pose-free methods (NopoSplat, STORM, DGGT, and Ours) on a sharp-turn sequence and a high-speed crossing sequence. NopoSplat rows show progressive blurring and misaligned projections. STORM rows exhibit sharpness degradation with visible ghosting artifacts. DGGT rows contain expanding black holes that grow larger as frames move further from context timestamps. The Ground4D rows maintain consistent structure, sharp vegetation detail, and complete spatial coverage throughout both sequences with no visible holes or blur.}
  \label{fig:consist_vis}
\end{figure*}

Table~\ref{tab:urban} evaluates Ground4D on the nuScenes urban benchmark against representative FFGS baselines. Fine-tuned on nuScenes, Ground4D achieves the highest PSNR of 27.23 and SSIM of 0.814. Urban scenes feature high sampling rates, lower frequency geometric structures and limited appearance variation, which reduces temporal conflicts among co-located Gaussians and allows intra-voxel aggregation to focus on geometric consistency.  Nevertheless, these results confirm that Ground4D could also extend effectively to urban driving scenes.

\subsection{Off-Road Reconstruction Visualization}
Figure~\ref{fig:contrast_vis} compares Ground4D against feedforward baselines across four scenarios: \textit{Day, Night, Snow,} and \textit{Forest}. MvSplat produces reasonable input-frame reconstruction but collapses at novel views due to depth estimation failure. DepthSplat leverages Depth Anything V2~\cite{yang2024depth} priors to stabilize depth, yet vibration noise in ground-truth off-road poses still induces geometric artifacts. NopoSplat, lacking dynamic modeling, suffers from ghosting and view misalignment under bumpy ego-motion, especially in the forest scene. DGGT's per-Gaussian temporal modeling neglects spatial occupancy, producing white holes in snow and semi-transparent artifacts in forest. STORM's optical flow proves counterproductive under off-road dynamics, distorting road surfaces at night and generating widespread artifacts in forest. Ground4D consistently delivers clean reconstructions across all conditions by confining temporal competition within local spatial voxels.

\subsection{Novel-View Synthesis under Sparse Inputs}
Figure~\ref{fig:consist_vis} visualizes two sequences of five consecutive novel-view frames depicting a sharp turn and a high-speed crossing, comparing four pose-free methods. NopoSplat exhibits severe blurring and misaligned Gaussian placements due to inaccurate pose and depth estimation. DGGT models temporal evolution per Gaussian but ignores spatial occupancy. As temporal distance from context frames grows, confidence decays uniformly, producing expanding structural holes. STORM's optical flow guidance remains sensitive to high-frequency vegetation and rapid illumination changes, degrading sharpness throughout. Ground4D suppresses both blurring and structural holes across all frames, confirming that intra-voxel spatial grounding maintains reconstruction fidelity even under large viewpoint changes with sparse temporal support.

\subsection{Ablation study}

\subsubsection{Voxel-Grounded Temporal Gaussian Aggregation.} Table~\ref{tab:modules_ablation} reveals a critical trade-off between temporal selectivity and spatial completeness. The naive introduction of Temporal Attention without local constraints, even when voxelized, fails to yield improvements and often degrades performance. We attribute this to the absence of localized competition among co-located Gaussians: each Gaussian scores its temporal relevance independently, without enforcing a dominant selection within each spatial region, causing spatial occupancy to collapse into structural voids (Figure~\ref{fig:teaser}).

While voxelization provides a geometric prior that acts as a spatial low-pass filter to anchor primitives, it lacks the discriminative capacity to resolve attribute ambiguities under rapid deformation. Our Intra-Voxel Normalization addresses this by re-parameterizing the competition via localized normalization. By confining the softmax within each voxel, temporal selectivity is enforced under occupancy-preserving constraints. This transforms the spatio-temporal interaction from competition to coordination: Temporal Attention identifies the optimal temporal state, while Intra-Voxel Normalization preserves geometric integrity. Such a dual-constraint mechanism effectively bridges high-frequency dynamics and stable geometric reconstruction in 4D.
\begin{table}[!h]
  \centering
  \caption{Ablation of Voxel-Grounded Temporal Gaussian Aggregation on ORAD-3D Dataset. TA and IN denote Temporal Attention and Intra-Voxel Normalization, respectively.}
  \label{tab:modules_ablation}
  \begin{tabular}{cccccc}
    \toprule
    Voxelize & TA & IN & PSNR $\uparrow$ & SSIM $\uparrow$ & LPIPS $\downarrow$ \\
    \midrule
    - & - & - &  22.45 & 0.62 & 0.34 \\
    - & \ding{52} & - & 21.70 & 0.61 & 0.33\\
    \ding{52} & - & - & 22.88 & 0.62 & 0.30\\
    \ding{52} & \ding{52} & - & 22.47 & 0.62 & 0.35\\
    \ding{52} & \ding{52} & \ding{52} & 23.89 & 0.64 & 0.23\\
    \bottomrule
  \end{tabular}
\end{table}

\subsubsection{Surface Normal Injection and Supervision.} Photometric supervision alone is insufficient to uniquely constrain Gaussian orientations in high-frequency off road scenes. We address this by introducing surface normal guidance through two complementary pathways. The loss $\mathcal{L}_{\text{norm}}^{\text{pred}}$ directly supervises the Normal Head with ground-truth normals, providing explicit geometric signals, while its intermediate features are injected into the Gaussian Head to condition scale and rotation predictions. This dual-path design couples explicit supervision with feature-level conditioning, enabling consistent orientation estimation and improving structural fidelity in photometrically ambiguous regions. Ablation results in Table~\ref{tab:sn_loss} further validate the effectiveness of this design.

\begin{table}[!h]
  \centering
  \caption{Ablation of Surface Normal Regularization.}
  \label{tab:sn_loss}
  \begin{tabular}{cccc}
    \toprule
    Method & PSNR $\uparrow$ & SSIM $\uparrow$ & LPIPS $\downarrow$ \\
    \midrule
    w/o normal & 23.45 & 0.62 & 0.27 \\
    ours & 23.89 & 0.64 & 0.23 \\
    \bottomrule
  \end{tabular}
\end{table}

\subsubsection{Voxel Size.} Table~\ref{tab:ablation_voxel} examines the effect of voxel size $\rho$. At the optimal $\rho = 0.002$, voxelization compresses the canonical Gaussian count from approximately 600K to 370K ($1.6\times$ reduction), merging redundant cross-frame observations while preserving spatial coverage and reducing rendering memory. Smaller voxels ($\rho = 0.001$) isolate co-located Gaussians into singleton voxels where temporal competition becomes trivial, while larger voxels ($\rho = 0.005$) over-merge spatially distinct surfaces, causing a severe 2.85~dB PSNR drop. This confirms that voxel scale must balance intra-voxel competition strength with spatial coherence.

\begin{table}[!h]
  \centering
  \caption{Ablation of Voxel Size on ORAD-3D Dataset.}
  \label{tab:ablation_voxel}
  \begin{tabular}{cccc}
    \toprule
    $\rho$ & PSNR $\uparrow$ & SSIM $\uparrow$ & LPIPS $\downarrow$ \\
    \midrule
    0.001 & 23.50 & 0.63& 0.24\\
    0.002 & 23.89 & 0.64 & 0.23\\
    0.003 & 23.71 & 0.63 & 0.25\\
    0.005 & 21.04 & 0.57 & 0.35\\
    \bottomrule
  \end{tabular}
\end{table}

\subsubsection{Context Input Frames.} Table~\ref{tab:multi_view} shows that increasing context frames initially improves reconstruction quality, as richer observations enhance temporal coverage and spatial support for voxel aggregation. The performance peaks at 10 frames and saturates thereafter. Further increasing the temporal window introduces observations from divergent deformation states, which exceed the reconciliation capacity of intra-voxel aggregation and degrade temporal coherence. This indicates that context size is inherently bounded by the model ability to resolve cross-time inconsistencies, with 10 frames providing an effective balance between spatial completeness and temporal coherence.

\begin{table}[!h]
  \centering
  \caption{Ablation of Context Frames Count.}
  \label{tab:multi_view}
  \begin{tabular}{ccccc}
    \toprule
    Context Frames & PSNR $\uparrow$ & SSIM $\uparrow$ & LPIPS $\downarrow$ \\
    \midrule
    4 & 23.89 & 0.64 & 0.23\\
    7 & 23.99 & 0.65 & 0.23\\
    10 & 24.05 & 0.65 & 0.22\\
    13 & 23.76 & 0.64 & 0.24\\
    16 & 23.65 & 0.63 & 0.25\\
    \bottomrule
  \end{tabular}
\end{table}

\section{Conclusion}
We have presented \textbf{Ground4D}, a spatially-grounded 4D feedforward framework for pose-free off-road scene reconstruction. The core insight is that confining temporal competition to local spatial voxels transforms selectivity and occupancy from competing objectives into mutually reinforcing ones, resolving the attribute conflicts that cause both blurring and structural holes. Built on this, voxel-grounded temporal gaussian aggregation and surface normal conditioning jointly address the compounding challenges of off-road scenes. Ground4D surpasses all feedforward baselines on ORAD-3D with up to 1.48\,dB PSNR gain and generalizes zero-shot to RELLIS-3D. 

\newpage
\bibliographystyle{ACM-Reference-Format}
\bibliography{ref}

\end{document}